\documentclass[twoside]{article}

\usepackage[preprint]{aistats2027}
\usepackage{amsmath}
\usepackage{booktabs}
\usepackage{graphicx}
\usepackage{url}
\usepackage{fancyvrb}
\usepackage{xcolor}
\usepackage[square,numbers]{natbib}
\usepackage{amsfonts}

\graphicspath{{figures/}}

\begin{document}

\runningtitle{Jev thinks ``I don't know'', but doesn't say it}
\runningauthor{Riccardo Porcedda}

\twocolumn[

\aistatstitle{Jev thinks ``I don't know'', but doesn't say it:\\Introducing Sys1Cal-v1 Dataset for Probability Calibration}

\aistatsauthor{Riccardo Porcedda}

\aistatsaddress{little-g.ai \\ Department of Excellence L'EMbeDS, Sant'Anna School of Advanced Studies, Italy \\ Department of Computer Science, University of Pisa, Italy} ]

\begin{abstract}
The appearance of Jev marked the era of System One Models, foundation models that return structured decisions with probability distributions rather than text. Aside from low cost and great speed, Jev's central promise is that these probabilities are calibrated: such claim is not backed by any public test and available external benchmarks evaluate confidence calibration, not whether every returned option probability has the right numerical meaning.
To tackle this issue, we introduce \textbf{Sys1Cal-v1}, a dataset of True/False questions about a proposition $A$ for which the exact probability $P(A)$ is known by construction. Each item is queried through the three Jev primitives - \texttt{Noul}, \texttt{Choice}, and \texttt{Score} - and evaluated by total variation distance from the ground-truth distribution, which can be used to estimate a soft accuracy of System One Models.\\
We showcase the utility of Sys1Cal-v1 as a benchmark dataset by evaluating Jev and SemIf, an open-source \texttt{Choice}-style baseline.

In this work, however, we focus even more deeply on Jev, by studying the calibration of its \texttt{Score} and \texttt{Choice} answers. In particular, we discover a peculiar behaviour that can be explained by assuming that Jev suppresses a third truth value, going beyond True and False. In other words, in \texttt{Choice} answers, $P(A)$ and $P(\neg A)$ are presented as if $P(A)+P(\neg A)=1$, while a term $P(U)\neq0$ is missing in the sum. Recovering $P(U)$ leads to an improvement of median soft accuracy in \texttt{Choice} answers from $0.771$ to $0.978$, suggesting that, even in binary decisions, Jev wants to answer with a third option:\\
\textit{``I don't know''}.
\end{abstract}

\section{Introduction} \label{sec:intro}

Probabilistic predictions are often consumed by downstream decision rules rather than used only to select the most likely class. Under Bayesian decision theory, a predictive distribution is combined with task-dependent costs or utilities to determine an optimal action; consequently, reliable class probabilities are important for cost-sensitive classification, autonomous decision systems, and settings in which a model may abstain or defer uncertain cases \cite{gneiting2007proper,silvafilho2023classifier,chow1970optimum,elyaniv2010foundations}. This motivates predictive interfaces in which probabilities are first-class outputs rather than auxiliary confidence scores. Jev was introduced by TypeSafe AI as a \emph{System One Model} for this setting: given an input state and typed questions, it returns structured probabilistic decisions rather than free-form text, through \emph{\texttt{Noul}} for binary judgments, \emph{\texttt{Choice}} for categorical decisions, and \emph{\texttt{Score}} for ordered scales \cite{typesafe2026jev}.

Leaving aside the great speed and low cost of this model, a question arises about probability calibration, since no test on available public datasets was performed.
The importance of assessing this calibration for process automatization is decision-theoretic: for actions $a$, outcomes $y$, and utilities $u(a,y)$, the optimal downstream action depends on the predictive distribution,
\[
  a^*(x)=\arg\max_a \sum_y P(y\mid x)u(a,y).
\]
Argmax accuracy evaluates only the case of a binary decision, for which the probability distribution collapses to a label and discards precisely the information that downstream decisions may need.
To give some examples:
\begin{itemize}
    \item classical probabilistic calibration asks whether events assigned probability $p$ occur with frequency $p$ \cite{dawid1982wellcalibrated};
    \item proper scoring rules such as the Brier and logarithmic scores reward truthful predictive distributions in expectation \cite{brier1950verification,gneiting2007proper};
    \item recent work motivates posterior-probability evaluation from Bayes decision theory \cite{ferrer2024posterior};
    \item modern neural-network work popularized confidence calibration and expected calibration error (ECE), where only the probability of the predicted class is evaluated \cite{guo2017calibration}. The JevBench dataset \cite{jevbench_github} belongs to this category.
\end{itemize}

Even if confidence calibration is an important test for System One models (we don't want class predictions to be underconfident/overconfident), this metric does not test whether each returned option probability has the right numerical meaning, whether equivalent states produce equivalent probabilities, or whether different Jev primitives expose compatible semantics (do probabilities from \texttt{Noul}, \texttt{Choice} and \texttt{Score} answers have the same meanings?).

We introduce \textbf{Sys1Cal-v1}\footnote{\url{https://github.com/little-g-ai/Sys1Cal-v1}} to isolate that missing question. Sys1Cal-v1 is a synthetic dataset in which the exact probability of a proposition is known by construction, evaluating the model's probability distribution rather than only empirical calibration against one realized label.

\subsection{Contributions}

We make four contributions.

\textbf{First}, we introduce \textbf{Sys1Cal-v1}, a benchmark dataset of True/False questions about a proposition $A$ for which $P(A)$ is known.

\textbf{Second}, we show how Sys1Cal-v1 can be used to evaluate System One models. We report \textit{Distributional Overlap} (OVL), i.e., soft accuracy, for Jev across \texttt{Noul}, \texttt{Choice}, and \texttt{Score}, and for SemIf \cite{semif_repo} as a \texttt{Choice}-style open-source baseline.

\textbf{Third}, we identify a systematic \texttt{Choice} miscalibration in Jev. While the same problem is present also in SemIf, in Jev there is a mapping that appears to solve the problem, suggesting that there exist an underlying hidden process defining how Jev assigns probabilities.

\textbf{Finally}, we show that this hidden process can be the presence of a third truth value, the ambiguity $U$, which goes beyond the concepts of True and False and identifies a state of uncertainty of the model. Estimating $U$ and using it to fix the probability distribution of \texttt{Choice} answers, increases the mean soft accuracy from 0.764 to 0.931.

\section{Sys1Cal-v1}

\subsection{Dataset Design}

Each Sys1Cal-v1 item begins with the definition of a proposition $A$. A random generator produces the probability
\[
  p^* = P(A), \text{ therefore } P(\neg A)=1-p^*.
\]
From these, we generate the states, which in Sys1Cal-v1 are of six types: explicit probabilities, frequencies from counts, compound probability, conditional probability, Bayes' rule, and sequential Bayesian updates. 
The aim behind these six types is checking to what difficulty level a System One model is able to perform calibrated decisions.
Furthermore, the same problem can be rendered in multiple equivalent forms: direct probability statements, counts, ratios, tables, prose, nested state, and distractor-augmented state. 

In total, the first version of Sys1Cal-v1 contains 365 rendered examples derived from 92 problems.
Here is a simplified example of how a Sys1Cal-v1 item would look like:
\small
\begin{verbatim}
{
  "family": "explicit_probability",
  "representation": "direct",
  "state": {
    "sufficient_statistics": {
      "p_A": 0.105263157895,
      "p_not_A": 0.894736842105
    }
  },
  "queries": {
    "noul": {
      "proposition": "Event A is true."
    },
    "choice": {
      "question": "What is the truth status of the 
      following proposition?",
      "proposition": "Event A is true.",
      "options": ["False", "True"]
    },
    "score": {
      "question": "To what degree is the 
      following proposition true?",
      "proposition": "Event A is true.",
      "levels": [
        "Completely false",
        "Very strongly false",
        "Strongly false",
        "Moderately false",
        "Slightly false",
        "Slightly true",
        "Moderately true",
        "Strongly true",
        "Very strongly true",
        "Completely true"
      ]
    }
  }}
\end{verbatim}

\normalsize

\begin{figure}[t]
\centerline{\includegraphics[width=\columnwidth]{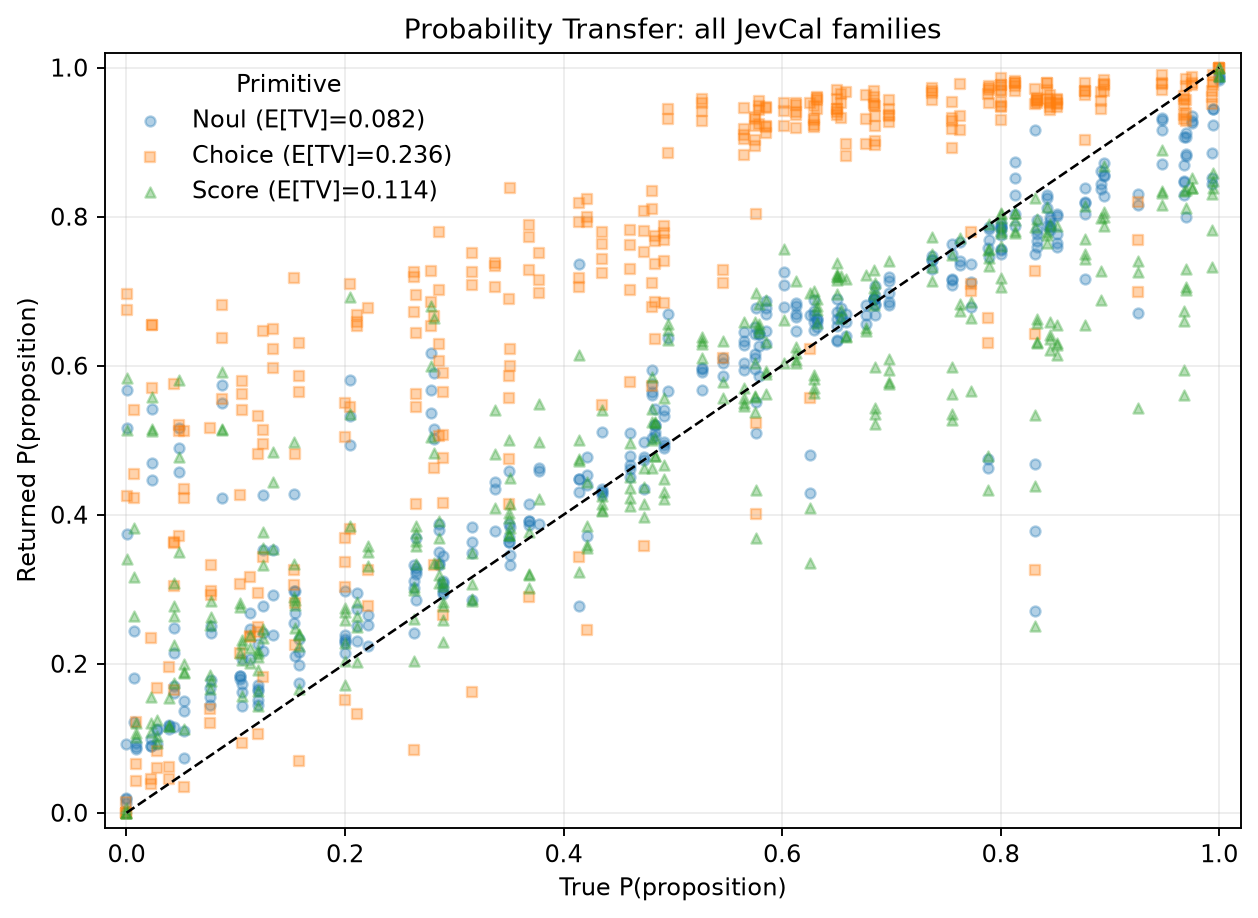}}
\caption{Probability transfer across all Sys1Cal-v1 families. \texttt{Noul} and \texttt{Score} expectation remain closer to the dashed identity line, while \texttt{Choice} is systematically shifted toward high \texttt{True} probabilities.}
\label{fig:prob-transfer}
\end{figure}

\subsection{Primitives and Projections}
We are going to discuss now what we expect from Jev's primitives and how answers and their probabilities are evaluated through Sys1Cal-v1.

\paragraph{Noul}
The type \texttt{Noul} is the simplest primitive: if $A$ is the prompt, it returns the estimated probabilities $P(A)$ and $P(\neg A)$, with $P(A)+P(\neg A)=1$.

\paragraph{Choice} This type returns a categorical distribution over defined criteria, i.e., possible answers to a given question. In our setting, the question is "What is the truth value of $A$?" and the criteria are \texttt{True} and \texttt{False}, therefore, we would expect the model to return $P(\texttt{True})=P(A)$ and $P(\texttt{False})=P(\neg A)$.

\paragraph{Score} This type returns an ordinal distribution over ordered, descriptive levels. So, for our purposes, while the question is the same as for \texttt{Choice}, here we define 10 levels corresponding to different grades of truth, from \texttt{Completely False} to \texttt{Completely True}. To each level $j$, we assign a numerical values $z_j=j/9$. The idea is to evaluate if the model is able to make fuzzy decisions.
Nonetheless, Sys1Cal-v1 only defines $P(\texttt{True})=P(A)$ and $P(\texttt{False})=P(\neg A)$, so, in order to evaluate probability calibration, we need a projection from numerical \texttt{Score} values into a binary \texttt{True}/\texttt{False} setting.

If $q_j$ is the score of level $j$ from the categorical distribution, and \texttt{Score} indeed represents a graded truth, we expect to get $P(A)$ from the \texttt{Score} \emph{expectation}
\begin{equation} \label{eq:score_exp}
    \mu_S=\sum_{j=0}^9 q_j z_j.
\end{equation}

\subsection{Metrics}

\begin{table}[t]
\caption{Mean TV and OVL (soft accuracy) by model-primitive. Jev-\texttt{Noul} and Jev-\texttt{Score} are well calibrated, while \texttt{Choice} answers show poorer performances on both Jev and SemIf}
\label{tab:main}
\begin{center}
\begin{tabular}{lrr}
\toprule
Output & Mean TV & Mean OVL \\
\midrule
Jev-\texttt{Choice} & 0.236 & 0.764 \\
Jev-\texttt{Noul} & 0.0817 & 0.918 \\
Jev-\texttt{Score} & 0.1139 & 0.886 \\
SemIf-\texttt{Choice} & 0.371 & 0.629 \\
\bottomrule
\end{tabular}
\end{center}
\end{table}

In Section \ref{sec:intro} we reported the main definitions of calibration. Here we set the metric used for System One models evaluation through Sys1Cal-v1: total variation distance.
Total variation distance is a statistical distance between two probability distributions (in this case, the true one and the empirical one returned by the model). For a random variable that can take only two values, this is
\[
  \mathrm{TV}(\hat p,p^*) = |\hat p-p^*|,
\]
with $\hat p$ being the estimated probability and $p^*$ the real one.
From this, we can also define the \emph{Distributional Overlap} (OVL), also known as the overlapping coefficient \cite{inman1989overlap}:
\[
  \mathrm{OVL}(\hat p,p^*)=1-|\hat p-p^*|.
\]
This quantity lies in $[0,1]$ and can be considered a \emph{soft accuracy}, since it generalizes the accuracy metric from deterministic one-hot targets to probabilistic targets. Such a metric couldn't be adopted in benchmarks where only true labels are known, discarding their probability.
But Sys1Cal-v1 is constructed specifically to make the target distribution available for this type of evaluation.

We also adopt $\mathrm{TV}(\hat p,p^*)$ to measure representation sensitivity: while a proposition can be expressed in multiple ways, the values of $P(A)$ remain the same, so we would expect from a good System One model to return the same probabilities, regardless of the proposition being presented as prose, table, counts, or a ratio.
We therefore group problems in Sys1Cal-v1 and measure the pairwise TV distance between equivalent renderings of the proposition.
The full summary appears in Appendix~\ref{app:repr}.

\begin{figure*}[t]
\centerline{\includegraphics[width=\textwidth]{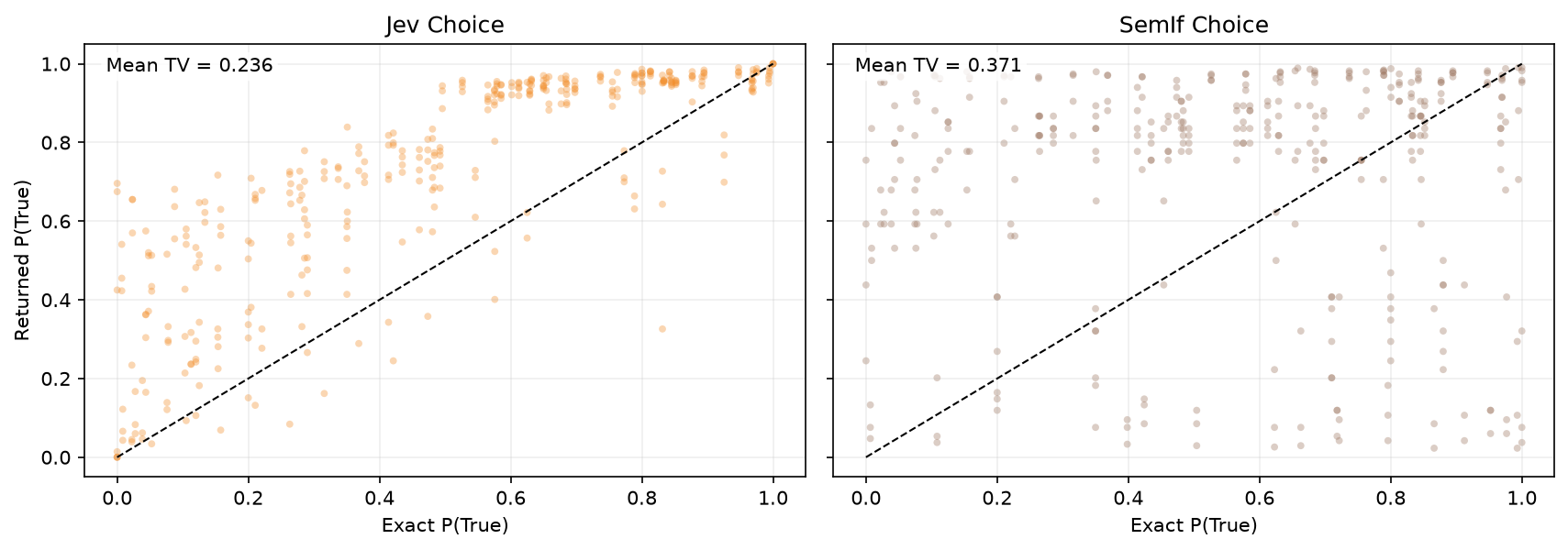}}
\caption{\texttt{Choice} probability transfer for Jev and SemIf. Mean TV is reported in each panel.}
\label{fig:choice-calibration}
\end{figure*}

\section{Evaluation of Probability Calibration}

Having defined the evaluation metrics, we proceed with the experiments on probability calibration using Sys1Cal-v1.
Since Jev and SemIf outputs are not deterministic, we input each Sys1Cal-v1 item 10 times and obtain the corresponding answers' probabilities $p_1,...,p_{10}$, from which we estimate $\hat p=\frac{1}{10}\sum_{i=1}^{10} p_i$. After this, we are able to compute $\mathrm{TV}(\hat p,p^*)$.

\subsection{Evaluating Primitives}
Instead of evaluating the overall probability calibration of the models, we want to study \texttt{Noul}, \texttt{Choice} and \texttt{Score} calibration separately. This is for two main reasons: having a fair comparison between Jev and SemIf (since the latter only produces \texttt{Choice} type answers) and studying in details the differences between Jev's primitives.
In Table~\ref{tab:main} we report the result.

While SemIf appears weaker in general, with an average OVL of 0.629, also Jev's \texttt{Choice} answers show worse calibration than \texttt{Noul} and \texttt{Score}. In Figure \ref{fig:prob-transfer} we provide a visual cue of this difference between primitives.
This is an interesting result, since, as we already highlighted, all the primitives are being evaluated on the same items and should estimate the same probabilities. In particular, it appears that Jev-\texttt{Choice} tends to return very high probabilities when $P(A)\geq0.5$, while being \emph{fuzzier} when $P(A)<0.5$. SemIf, on the other hand, appears to be generally miscalibrated (see Figure \ref{fig:choice-calibration}).


\begin{figure*}[t]
\centerline{\includegraphics[width=\textwidth]{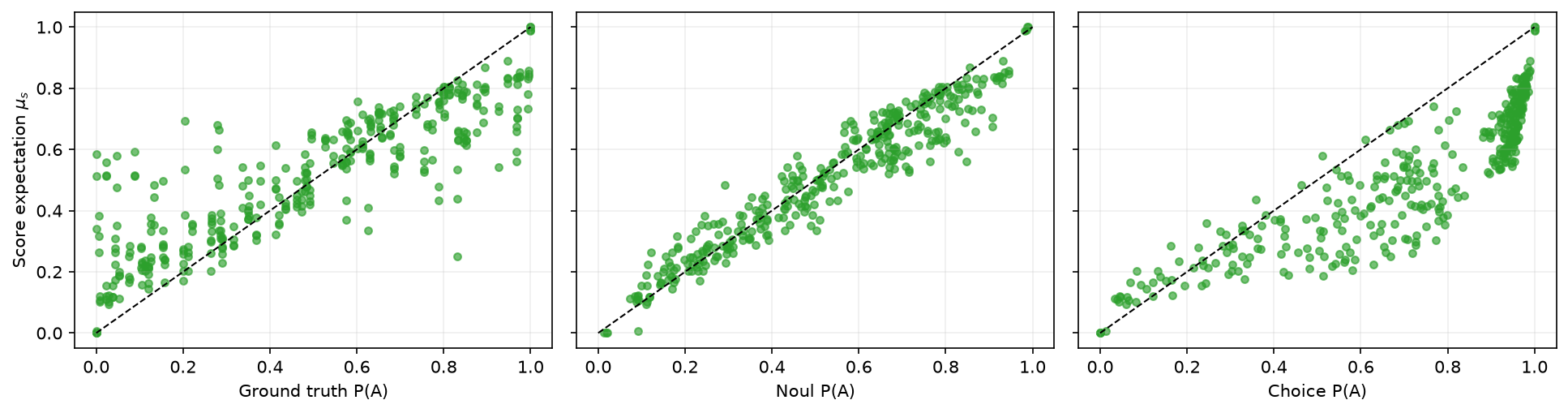}}
\caption{\texttt{Score} expectations compared with ground truth, \texttt{Noul}, and \texttt{Choice} probabilities. \texttt{Score} expectations tracks \texttt{Noul} more tightly than it tracks \texttt{Choice}, which shows a peculiar shape that suggests the existence of a map that could fix the calibration.}
\label{fig:score-connections}
\end{figure*}

Jev-\texttt{Noul} and Jev-\texttt{Score} show a similar calibration, obtaining an average OVL of $0.918$ and $0.886$, respectively, showing that the \texttt{Score} expectation defined in Equation \ref{eq:score_exp} correctly recovers the probabilities returned by Jev-\texttt{Noul}.
This also motivates us to further study the correlation between \texttt{Score} expectations and the probabilities returned by Jev-\texttt{Noul} and Jev-\texttt{Choice}.

\subsection{A hint from Score expectations}
In Figure \ref{fig:score-connections} we plot \texttt{Score} expectations against ground truth, \texttt{Noul}, and \texttt{Choice} probabilities. From the first plot, we can assess the goodness of the calibration of \texttt{Score} answers, and, in the second plot, we show the agreement between \texttt{Score} expectations and \texttt{Noul}.
But, more interestingly, the last plot shows that \texttt{Choice} answers are not merely miscalibrated: it appears that there exists a non-linear map from \texttt{Score} expectations that could fix the probability calibration.

This raises the following question: \textit{what kind of latent behavior would distort \texttt{Choice} probabilities in such a systematic way, while leaving \texttt{Noul} and \texttt{Score} expectation well correlated and much closer to the ground truth?}

\section{Going Beyond True and False}

In Sys1Cal-v1, \texttt{Score} answers return an ordinal distribution
ranging from \texttt{Completely False} to \texttt{Completely True}.
We summarize this distribution through its expectation $\mu_S$, as
defined in Equation~\ref{eq:score_exp}. A direct binary interpretation
would therefore associate $\mu_S$ with the probability of
\texttt{True} and $1-\mu_S$ with the probability of \texttt{False}.

As we have shown, this projection is linearly related to
\texttt{Noul} probabilities. Its relation with binary
\texttt{Choice} probabilities, however, is non-linear.
This suggests that the transformation from \texttt{Score} to
\texttt{Choice} may discard information that cannot be represented by
a direct True/False projection.

We therefore introduce a latent three-component probability
distribution
\begin{equation*}
    \pi_T+\pi_U+\pi_F=1,
\end{equation*}
where $\pi_T$, $\pi_U$, and $\pi_F$ denote, respectively, the latent
probabilities associated with \texttt{True}, \texttt{Uncertain}, and
\texttt{False}.

If \texttt{Choice} excludes the uncertain component and renormalizes
the remaining two probabilities, then
\begin{align}
P_{\texttt{Choice}}(\texttt{True})
    &=
    \frac{\pi_T}{\pi_T+\pi_F}
    \\
    &=
    \frac{\pi_T}{1-\pi_U},
\label{eq:choice-renorm}
\end{align}
and analogously
\[
P_{\texttt{Choice}}(\texttt{False})
=
\frac{\pi_F}{1-\pi_U}.
\]

We infer this latent representation from the \texttt{Score}
expectation $\mu_S$. In particular, we take the \texttt{Score}
expectation as the latent probability already assigned to
\texttt{True},
\begin{equation}
    \widehat{\pi}_T=\mu_S,
\end{equation}
and allow part of the remaining probability $1-\mu_S$ to represent
uncertainty.
The reason why we assume this is that \texttt{Score}
expectations and \texttt{Noul} answers (which are strongly correlated) appear to be well calibrated, so no relevant uncertainty appears to be present in $\mu_S$.

We model this latent uncertainty probability as
\begin{equation*}
    \widehat{\pi}_U(\mu_S)
    =
    \mu_S^\alpha(1-\mu_S)^\beta,
\label{eq:latent-uncertainty}
\end{equation*}
with $\alpha>0$ and $\beta\geq1$. These constraints guarantee that
\[
0\leq \widehat{\pi}_U(\mu_S)\leq 1-\mu_S.
\]
The remaining probability is assigned to \texttt{False},
\begin{equation*}
    \widehat{\pi}_F
    =
    1-\widehat{\pi}_T-\widehat{\pi}_U.
\end{equation*}

The corresponding prediction of the binary \texttt{Choice}
probability is therefore
\begin{equation}
\widehat{P}_{\texttt{Choice}}(\texttt{True}\mid\mu_S)
=
\frac{\mu_S}
{1-\mu_S^\alpha(1-\mu_S)^\beta}.
\label{eq:score-choice-model}
\end{equation}

To test whether a non-zero uncertainty component is supported by the data,
we introduce a scale parameter
\[
\widehat{\pi}_U(\mu_S)
=
\lambda\,\mu_S^\alpha(1-\mu_S)^\beta,
\qquad 0\leq\lambda\leq1,
\]
so that the no-uncertainty hypothesis is $H_0:\lambda=0$.
We estimate both a global $\lambda$ and problem-level values $\lambda_j$
across the 92 latent problems; Table~\ref{tab:ambiguity-tests} reports
the corresponding estimates and tests.

\begin{figure}[t]
\centerline{\includegraphics[width=\columnwidth]{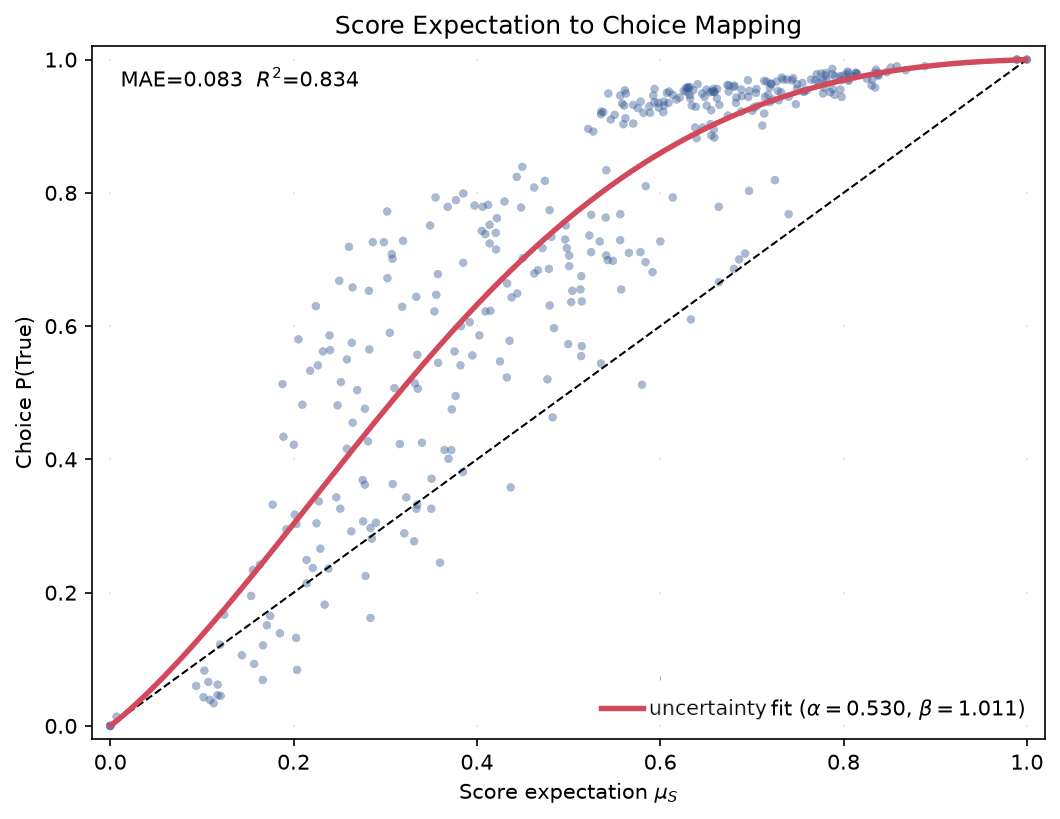}}
\caption{
\texttt{Score}-to-\texttt{Choice} mapping induced by the fitted latent
uncertainty model. Introducing a latent uncertainty probability of the
form $\mu_S^\alpha(1-\mu_S)^\beta$ captures much of the systematic
non-linearity between \texttt{Score} expectations and binary
\texttt{Choice} probabilities.
}
\label{fig:noul-choice}
\end{figure}

Fitting Eq. \ref{eq:score-choice-model} to Sys1Cal-v1 yields
\[
\alpha=0.530,
\qquad
\beta=1.011,
\]
with $R^2=0.834$ for the resulting prediction of
\texttt{Choice} probabilities (see Figure \ref{fig:noul-choice}).

More importantly, introducing this
component improves the prediction of binary \texttt{Choice}
probabilities: the mean reduction in absolute error is $0.110$, with
a 95\% confidence interval that excludes zero.

While our definition of uncertainty need not to be an exact discovery of how Jev encodes truth values, these tests show that \textbf{Jev-\texttt{Choice} may go beyond a True/False-only setting}.

\section{Uncertainty as a Useful Signal for Choice}

The uncertainty model is not only a post-hoc explanation of the Score--Choice discrepancy. It yields two operationally different objects. The first is a calibrated binary \texttt{Choice} probability, useful when downstream systems require the original \texttt{True/False} setting. The second is a three-status representation, useful when a system encodes uncertainty and act conditionally on it.

Let
\[
  f(\mu)
  =
  \frac{\mu}{1-\lambda\mu^\alpha(1-\mu)^\beta}
\]
be the fitted Score-to-Choice distortion map. If \texttt{Choice} behaves like a binary projection of a richer state, then a raw \texttt{Choice} probability $p_C$ can be corrected by applying the inverse map
\[
  \tilde p_C=f^{-1}(p_C).
\]
This gives an ordinary binary probability estimate, so it can be evaluated with the same distributional overlap used for raw Choice:
\[
  \mathrm{OVL}_{\mathrm{corr}}
  =
  1-\left|\tilde p_C-p^*\right|.
\]
The correction substantially improves binary probability recovery (see Table \ref{tab:choice-uncertainty-ovl}). Raw Choice has mean OVL $0.764$ and median OVL $0.771$, whereas inverse uncertainty calibration raises these values to $0.880$ and $0.903$, respectively. Thus the uncertainty model is not merely descriptive: when inverted, it acts as a practical post-hoc calibration layer for Choice probabilities.

\begin{table}[t]
  \caption{Tests for the scaled Score--Uncertainty model.
  The null hypothesis is $H_0:\lambda=0$.
  The global $\hat\lambda$ row reports the fitted scale and a bootstrap
  confidence interval over latent problems. The $\lambda_j$ tests whether the fitted uncertainty is positive across the latent problems.}
  \label{tab:ambiguity-tests}
  \begin{center}
  \begin{tabular}{lrrr}
  \toprule
  Quantity & Estimate & 95\% CI & $p$-value \\
  \midrule
  Global $\hat\lambda$ & 0.978 & [0.789, 1.000] & -- \\
  Latent $\lambda_j$ & 0.821 & [0.760, 0.882] & $2.94{\times}10^{-44}$ \\
  \bottomrule
  \end{tabular}
  \end{center}
\end{table}

The second use keeps the inferred uncertainty mass instead of forcing it back into a binary probability. The fitted model induces
\[
  T=\mu_S,\qquad
  U=\lambda\mu_S^\alpha(1-\mu_S)^\beta,\qquad
  F=1-T-U.
\]
This three-status representation does not assert a single value for $P(A)$. Instead, it defines an interval of compatible truth probabilities,
\begin{equation}\label{eq:credal-interval}
  P(A)\in[T,T+U].
\end{equation}
We therefore evaluate it with
\[
  \mathrm{OVL}_{TUF}
  =
  1-d\bigl(p^*,[T,T+U]\bigr),
\]
where $d$ is the absolute distance from $p^*$ to the interval. This quantity is not directly identical to binary OVL: a wider interval is more permissive. For this reason, $\mathrm{OVL}_{TUF}$ must be reported together with the width $U$. In our evaluation, the T/U/F interval reaches mean OVL $0.931$ and median OVL $0.978$, with mean uncertainty width $0.292$.

These two uses support different deployment patterns. Corrected Choice is appropriate when an application needs a single calibrated binary probability for thresholding, ranking, expected-utility decisions, or risk scoring. It preserves the original Choice interface while reducing its probability distortion. The T/U/F representation is appropriate when the system can make uncertainty-aware decisions: act automatically when $U$ is small, defer when $U$ is large, request more evidence when the interval crosses a decision threshold, or route the item to a slower model or human reviewer. This connects JevCal to selective prediction and abstention, where uncertainty is valuable precisely because it identifies cases in which an automatic binary decision should be treated cautiously \cite{xin2021art,kadavath2022language}.

The practical conclusion is that the Choice error is recoverable in two ways. If a binary answer is required, inverse uncertainty calibration turns distorted Choice probabilities into much better probability estimates. If the interface can expose richer state, the inferred T/U/F representation provides a more informative object: not only an estimate of truth support, but also a measure of how much probability mass was unresolved by the forced binary projection.

\begin{table}[t]
\caption{\texttt{Choice} OVL under three evaluation modes. The T/U/F score evaluates interval compatibility and should be interpreted together with the mean uncertainty width $\mathbb{E}[U]$.}
\label{tab:choice-uncertainty-ovl}
\begin{center}
\begin{tabular}{lrr}
\toprule
Setting & Mean OVL & Med. OVL \\
\midrule
Raw \texttt{Choice} & 0.764 & 0.771\\
Calibrated \texttt{Choice} & 0.880 & 0.903\\
T/U/F & 0.931 & 0.978 \\
\bottomrule
\end{tabular}
\end{center}
\end{table}

\paragraph{Practical decision example.}

Consider an automated agent deciding whether a proposition $A$ is true
enough to trigger an action, for example whether a transaction should be
approved automatically or whether an e-mail should be marked as spam.
The agent has three actions:
\begin{align*}
a_T&=\text{act as if }A\text{ is true},\\
a_F&=\text{act as if }A\text{ is false},\\
a_D&=\text{defer}.
\end{align*}
A correct committed decision has zero loss, an incorrect committed
decision has cost $C_{\mathrm{err}}=100$, and deferral to a slower or
human procedure has cost $C_D=45$.

Suppose the raw \texttt{Choice} output is
\[
P_{\texttt{Choice}}(A)=0.632.
\]
Using this probability directly, the expected losses are
\begin{align*}
R(a_T)&=100(1-0.632)=36.8,\\
R(a_F)&=100(0.632)=63.2,\\
R(a_D)&=45.
\end{align*}
Thus raw \texttt{Choice} recommends committing to $a_T$.

Now apply the inverse uncertainty calibration map. For this example,
\[
\tilde p_C=f^{-1}(0.632)\simeq 0.400.
\]
The calibrated binary probability reverses the preferred committed
decision. The expected losses become
\begin{align*}
R(a_T)&=100(1-0.400)=60.0,\\
R(a_F)&=100(0.400)=40.0,\\
R(a_D)&=45.
\end{align*}
Thus calibrated \texttt{Choice} recommends committing to $a_F$.

Finally, keep the inferred three-status representation:
\[
(\widehat{\pi}_T,\widehat{\pi}_U,\widehat{\pi}_F)
=
(0.400,0.367,0.233).
\]
Renormalizing the committed components gives
\[
\frac{\widehat{\pi}_T}
     {\widehat{\pi}_T+\widehat{\pi}_F}
=
\frac{0.400}{0.400+0.233}
\simeq 0.632,
\]
so the T/U/F representation explains the raw \texttt{Choice} output as
a forced binary projection. However, it also preserves the unresolved
mass. By Equation~\ref{eq:credal-interval}, it induces the compatible
probability interval
\[
P(A)\in[0.400,0.767].
\]
Under a robust $\Gamma$-minimax criterion
\cite{berger1985statistical,walley1991statistical}, the worst-case
losses are
\begin{align*}
\overline R(a_T)&=100(1-0.400)=60.0,\\
\overline R(a_F)&=100(0.767)=76.7,\\
\overline R(a_D)&=45.
\end{align*}
The robust action is therefore $a_D$.

\begin{table}[t]
\caption{Decision induced by the three Choice-derived settings in a
costly automation example. The same raw \texttt{Choice} output leads to
three different actions: raw \texttt{Choice} commits to \texttt{True},
calibrated \texttt{Choice} commits to \texttt{False}, and T/U/F defers
because the unresolved mass makes either committed action too risky.}
\label{tab:choice-decision-example}
\begin{center}
\begin{tabular}{lll}
\toprule
Setting & Criterion & Action \\
\midrule
Raw \texttt{Choice} &
Bayes risk &
\textbf{Act True} \\
Calibrated \texttt{Choice} &
Bayes risk &
\textbf{Act False} \\
T/U/F &
Robust risk &
\textbf{Defer} \\
\bottomrule
\end{tabular}
\end{center}
\end{table}

\section{Scope and Limitations}

Sys1Cal-v1 is designed to isolate probability semantics, not to measure
broad natural-language competence. Its synthetic construction is a
strength: each item has an exact pointwise probability, so model outputs
can be compared directly with the target distribution. This is precisely
what makes distributional overlap meaningful in our setting. At the same
time, the benchmark has limited ecological coverage. The current release
uses controlled probability families and templated renderings; future
versions should include richer linguistic variation, adversarial
paraphrases, domain-specific decision problems, and independently
validated natural-language formulations.

Our analysis of \texttt{Score} also relies on a one-dimensional
projection: the expectation of a 10-level ordered distribution. This
projection is natural for truth-like scores, but it may not exhaust the
information contained in the full \texttt{Score} distribution. Other
summaries, or direct evaluation of the full ordered distribution, may
expose additional structure.

Finally, the uncertainty model should be interpreted as an observational
account of the relation between \texttt{Score} and \texttt{Choice}, not
as a causal claim about Jev's internal implementation. The results show
that \texttt{Choice} behaves as if a richer state were being collapsed
into a forced binary output, and that the inferred unresolved mass is
useful for calibration and decision making. They indicate, but they don't prove, that Jev
explicitly represents a hidden third truth value internally.

\section{Conclusion}

We introduced \textbf{Sys1Cal-v1}, a benchmark for evaluating whether the
probabilities returned by System One models have the intended numerical
meaning. Unlike confidence-calibration benchmarks, Sys1Cal-v1 provides
the exact probability of each proposition by construction and evaluates
the returned distribution directly. This makes it possible to test
pointwise probability recovery, compare probabilistic semantics across
different primitives, and measure representation sensitivity across
equivalent formulations of the same latent problem.

Our evaluation shows that Jev's primitives do not expose probability in
the same way. \texttt{Noul} and \texttt{Score} are substantially better
aligned with the ground-truth probabilities than \texttt{Choice}. The
benchmark also reveals representation sensitivity: equivalent
formulations of the same probability problem can induce different
outputs, with \texttt{Choice} less invariant than \texttt{Noul} and
\texttt{Score}.

We then showed that the \texttt{Choice} distortion can be explained by an uncertainty model estimated from \texttt{Score} answers.
Inverting this map gives a practical post-hoc calibration layer for
binary \texttt{Choice}: median binary OVL improves from $0.771$ for raw
\texttt{Choice} to $0.903$ after correction. Keeping the inferred
uncertainty instead yields a T/U/F interval representation with
median interval OVL $0.971$. The latter
score is not directly interchangeable with binary OVL, but it captures a
different and useful object: compatibility with a range of truth
probabilities induced by an uncertain component.

This distinction matters for downstream automation. We showed with a practical example how the outcome of a selective decision problem can be different using raw
\texttt{Choice}, calibrated \texttt{Choice} or T/U/F.

The broader lesson is that evaluation of structured decision models
should not stop at argmax accuracy or top-label confidence calibration.
If a model returns probabilities for use in automated decisions, the
probabilities themselves are the object being promised. Sys1Cal-v1 makes
that promise testable, and shows that probability semantics can differ
substantially across interfaces even within the same model.

\bibliography{main}

\clearpage
\appendix
\thispagestyle{empty}

\onecolumn

\section{Representation Sensitivity}
\label{app:repr}

Representation sensitivity measures semantic invariance: if two prompts encode the same probability problem, a calibrated probabilistic interface should return the same distribution up to noise. Each latent problem is rendered in multiple equivalent forms; we compute pairwise TV across renderings after aggregating repeats, and also record the maximum spread within each latent group. Lower values mean greater invariance.

The results show a consistent ordering. Jev \texttt{Noul} is the most stable primitive, with mean pairwise TV 0.044. \texttt{Score} expectation is less stable than \texttt{Noul} but still substantially more stable than \texttt{Choice}. Jev \texttt{Choice} has more than twice \texttt{Noul}'s mean pairwise variation, while SemIf \texttt{Choice} is the least stable model in this comparison. This matters because representation sensitivity can hide behind aggregate calibration: a model may achieve a reasonable average error while still changing its probabilities when the same state is written in a different but equivalent form.

\begin{table*}[h]
\caption{Examples of representation types in Sys1Cal-v1. Each row shows one rendered state format used to express an exact latent probability problem. The proposition and gold distribution are shared by the three primitives Noul, Choice, and Score.}
\label{tab:representation-examples}
\begin{center}
\scriptsize
\begin{tabular}{llp{0.4\textwidth}p{0.15\textwidth}}
\toprule
Representation & Family & State & Proposition / gold \\
\midrule
Direct &
Explicit probability &
\texttt{sufficient\_statistics:} $p_A=0,\;p_{\neg A}=1$ &
``Event A is true.'' \newline
$P(\mathrm{True})=0$ \\\\
\hline
Ratio &
Explicit probability &
\texttt{probability\_ratio\_A\_to\_not\_A: 0:1} &
``Event A is true.'' \newline
$P(\mathrm{True})=0$ \\\\
\hline
Prose &
Explicit probability &
Natural-language rendering: ``The state fully determines the target distribution. The probability of $A$ is 0, and the remaining probability belongs to the other outcome.'' &
``Event A is true.'' \newline
$P(\mathrm{True})=0$ \\
\hline
Distractor &
Explicit probability &
Same sufficient statistics as the direct form, plus irrelevant metadata &
``Event A is true.'' \newline
$P(\mathrm{True})=0$ \\\\
\hline
Counts &
Frequency &
\texttt{counts:} zor $=800$, nif $=200$; sampling is uniform over objects. &
``The sampled object is a zor.'' \newline
$P(\mathrm{True})=0.8$ \\\\
\hline
Scaled counts &
Frequency &
Equivalent count representation with all counts scaled: zor $=80000$, nif $=20000$. &
``The sampled object is a zor.'' \newline
$P(\mathrm{True})=0.8$ \\\\
\hline
Table &
Frequency &
A table with one row per label: zor has count 40, nif has count 10. &
``The sampled object is a zor.'' \newline
$P(\mathrm{True})=0.8$ \\\\
\hline
Nested state &
Conditional probability &
Nested contingency state with counts:
$A\wedge B=74$, $A\wedge \neg B=48$, $\neg A\wedge B=79$, $\neg A\wedge \neg B=55$;
query is $P(A\mid B)$. &
``Event A is true.'' \newline
$P(\mathrm{True})=0.484$ \\
\bottomrule
\end{tabular}
\end{center}
\end{table*}

\clearpage

\begin{table*}[t]
\caption{Representation sensitivity for Jev and SemIf. Mean pairwise TV is the average distance between equivalent renderings of the same latent problem.}
\label{tab:repr}
\begin{center}
\footnotesize
\begin{tabular}{@{}lrrrr@{}}
\toprule
Output & Mean pairwise TV & Mean max TV \\
\midrule
\texttt{Choice} & 0.0998 & 0.181 \\
\texttt{Noul} & 0.0441 & 0.0787 \\
\texttt{Score} exp. & 0.0656 & 0.116\\
SemIf \texttt{Choice} & 0.117 & 0.194 \\
\bottomrule
\end{tabular}
\end{center}
\end{table*}

\begin{figure*}[h]
\centerline{\includegraphics[width=0.7\columnwidth]{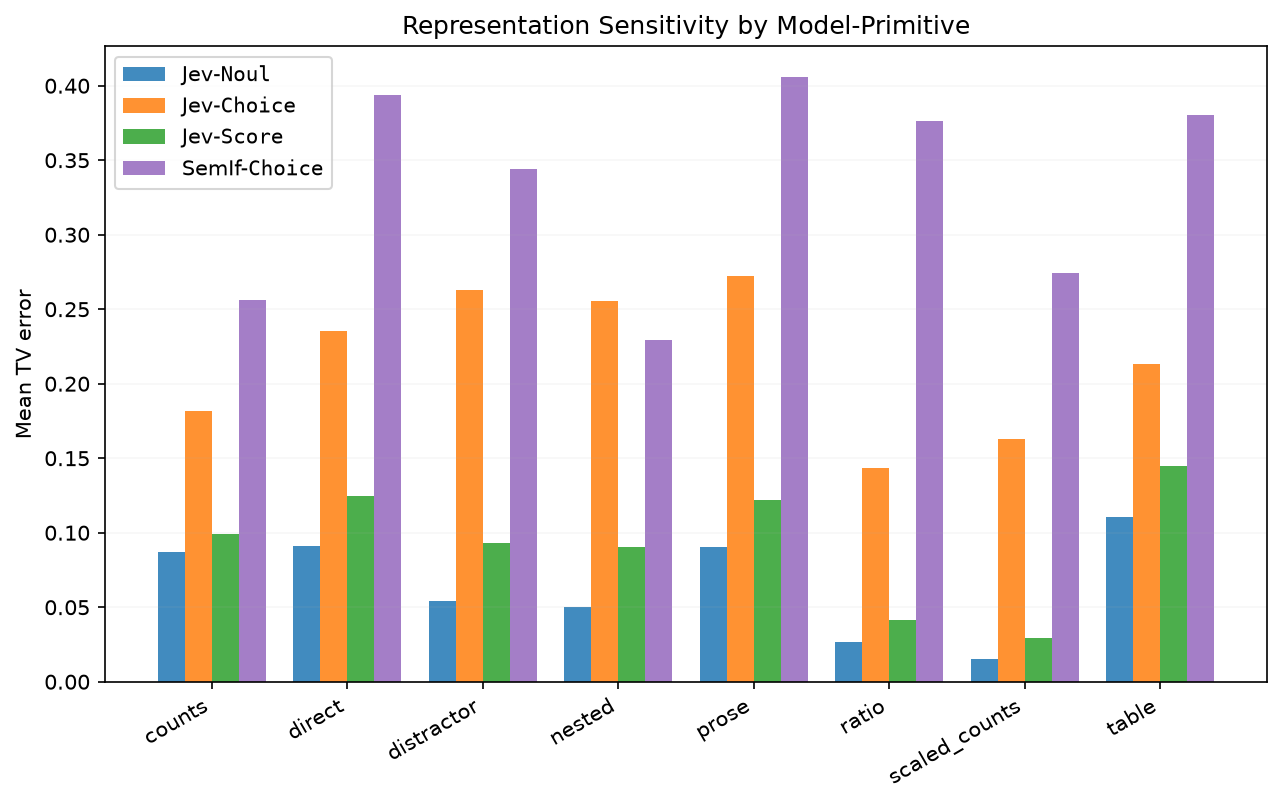}}
\caption{Representation sensitivity by primitive. \texttt{Choice} and SemIf are more sensitive to equivalent renderings than \texttt{Noul} or \texttt{Score} expectation.}
\label{fig:repr-sensitivity}
\end{figure*}

\end{document}